# Learning Invariant Representations with Local Transformations

Kihyuk Sohn                                                                                       KIHYUKS@UMICH.EDU
Honglak Lee                                                                              HONGLAK@EECS.UMICH.EDU
Dept. of Electrical Engineering and Computer Science, University of Michigan, Ann Arbor, MI 48109, USA

## Abstract

Learning invariant representations is an important problem in machine learning and pattern recognition. In this paper, we present a novel framework of transformation-invariant feature learning by incorporating linear transformations into the feature learning algorithms. For example, we present the transformation-invariant restricted Boltzmann machine that compactly represents data by its weights and their transformations, which achieves invariance of the feature representation via probabilistic max pooling. In addition, we show that our transformation-invariant feature learning framework can also be extended to other unsupervised learning methods, such as autoencoders or sparse coding. We evaluate our method on several image classification benchmark datasets, such as MNIST variations, CIFAR-10, and STL-10, and show competitive or superior classification performance when compared to the state-of-the-art. Furthermore, our method achieves state-of-the-art performance on phone classification tasks with the TIMIT dataset, which demonstrates wide applicability of our proposed algorithms to other domains.

## 1. Introduction

In recent years, unsupervised feature learning algorithms have emerged as promising tools for learning representations from data (Hinton et al., 2006; Bengio et al., 2007; Ranzato et al., 2006). In particular, it is an important problem to learn invariant representations that are robust to variability in high-dimensional data (e.g., images, speech, etc.) since they will enable machine learning systems to achieve good generalization performance while using a small number of labeled training examples. In this context, several feature learning algorithms have been proposed to learn invariant representations for specific transformations by using customized approaches. For example, convolutional feature learning methods (Lee et al., 2011; Kavukcuoglu et al., 2010; Zeiler et al., 2010) can achieve shift-invariance by exploiting convolution operators. As another example, the denoising autoencoder (Vincent et al., 2008) can learn features that are robust to the input noise by trying to reconstruct the original data from the hidden representation of the perturbed data. However, learning invariant representations with respect to general types of transformations is still a challenging problem.

In this paper, we present a novel framework of transformation-invariant feature learning. We focus on local transformations (e.g., small amounts of translation, rotation, and scaling in images), which can be approximated as linear transformations, and incorporate linear transformation operators into the feature learning algorithms. For example, we present the transformation-invariant restricted Boltzmann machine, which is a generative model that represents input data as a combination of transformed weights. In this case, a transformation-invariant feature representation is obtained via probabilistic max pooling of the hidden units over the set of transformations. In addition, we show extensions of our transformation-invariant feature learning framework to other unsupervised feature learning algorithms, such as autoencoders or sparse coding.

In our experiments, we evaluate our method on the variations of the MNIST dataset and show that our algorithm can significantly outperform the baseline restricted Boltzmann machine when underlying transformations in the data are well-matched to those considered in the model. Furthermore, our method can learn features that are much more robust to the wide range of local transformations, which results in highly competitive performance in visual recognition tasks on CIFAR-10 (Krizhevsky, 2009) and STL-10 (Coates et al., 2011) datasets. In addition, our method also achieves state-of-the-art performance on phone classification tasks with the TIMIT dataset, which demonstrates wide applicability of our proposed algorithms to other domains.

The rest of the paper is organized as follows. We provide the preliminaries in Section 2, and in Section 3, we introduce our proposed transformation-invariant feature learning algorithms. In Section 4, we review the previous work on invariant feature learning. Then, in Section 5, we report the experimental results on several datasets. Section 6 concludes the paper.

## 2. Preliminaries

In this paper, we present a general framework for learning locally-invariant features using transformations. For presentation, we will use the restricted Boltzmann machine (RBM) as the main example.[1] We first describe the RBM below, followed by its novel extension (Section 3).

The restricted Boltzmann machine is a bipartite undirected graphical model that is composed of visible and hidden layers. Assuming binary-valued visible and hidden units, the energy function and the joint probability distribution are given as follows:[2]

$$E(\mathbf{v}, \mathbf{h}) = -\mathbf{v}^T \mathbf{W} \mathbf{h} - \mathbf{b}^T \mathbf{h} - \mathbf{c}^T \mathbf{v}, \quad (1)$$
$$P(\mathbf{v}, \mathbf{h}) = \frac{1}{Z} \exp\left(-E(\mathbf{v}, \mathbf{h})\right) \quad (2)$$

where $\mathbf{v} \in \{0,1\}^D$ are binary visible units, $\mathbf{h} \in \{0,1\}^K$ are binary hidden units, and $\mathbf{W} \in \mathbb{R}^{D \times K}$, $\mathbf{b} \in \mathbb{R}^K$, and $\mathbf{c} \in \mathbb{R}^D$ are weights, hidden biases, and visible biases, respectively. $Z$ is a normalization factor that depends on the parameters $\{\mathbf{W}, \mathbf{b}, \mathbf{c}\}$. Since RBMs have no intra-layer connectivity, exact inference is tractable and block Gibbs sampling can be done efficiently using the following conditional probabilities:

$$p(h_j = 1 | \mathbf{v}) = \text{sigmoid}(\sum_i v_i W_{ij} + b_j) \quad (3)$$
$$p(v_i = 1 | \mathbf{h}) = \text{sigmoid}(\sum_j h_j W_{ij} + c_i) \quad (4)$$

where $\text{sigmoid}(x) = \frac{1}{1+\exp(-x)}$. We train the RBM parameters by minimizing the negative log-likelihood via stochastic gradient descent. Although computing the exact gradient is intractable, we can approximate it using contrastive divergence (Hinton, 2002).

## 3. Learning Transformation-Invariant Feature Representations

### 3.1. Transformation-invariant RBM

In this section, we formulate a novel feature learning framework that can learn invariance to a set of linear transformations based on the RBM. We begin the section with describing the transformation operator.

The transformation operator is defined as a mapping $T : \mathbb{R}^{D_1} \to \mathbb{R}^{D_2}$ that maps $D_1$-dimensional input vectors into $D_2$-dimensional output vectors ($D_1 \geq D_2$). In our case, we assume a linear transformation matrix $T \in \mathbb{R}^{D_2 \times D_1}$, i.e., each coordinate of the output vector is represented as a linear combination of the input coordinates.

With this notation, we formulate the *transformation-invariant restricted Boltzmann machine* (TIRBM) that can learn invariance to a set of transformations. Specifically, for a given set of transformation matrices $T_s$ ($s = 1, \cdots, S$), the energy function of TIRBM is defined as follows:

$$E(\mathbf{v}, \mathbf{H}) = \quad (5)$$
$$-\sum_{j=1}^{K}\sum_{s=1}^{S}(T_s \mathbf{v})^T \mathbf{w}_j h_{j,s} - \sum_{j=1}^{K}\sum_{s=1}^{S} b_{j,s} h_{j,s} - \mathbf{c}^T \mathbf{v}$$
$$\text{s.t.} \sum_{s=1}^{S} h_{j,s} \leq 1, h_{j,s} \in \{0,1\}, j = 1, \cdots, K, \quad (6)$$

where $\mathbf{v}$ are $D_1$-dimensional visible units, and $\mathbf{w}_j$ are $D_2$-dimensional (filter) weights corresponding to the $j$-th hidden unit. The hidden units are represented as a matrix $\mathbf{H} \in \{0,1\}^{K \times S}$ with $h_{j,s}$ as its $(j,s)$-th entry. In addition, we denote $z_j = \sum_{s=1}^{S} h_{j,s}, z_j \in \{0,1\}$ as a pooled hidden unit over the transformations.

In Equation (6), we impose a softmax constraint on hidden units so that at most one unit is activated at each row of $\mathbf{H}$. This probabilistic max pooling[3] allows us to obtain a feature representation invariant to linear transformations. More precisely, suppose that the input $\mathbf{v}_1$ matches the filter $\mathbf{w}_j$. Given another input $\mathbf{v}_2$ that is a transformed version of $\mathbf{v}_1$, the TIRBM will try to find a transformation matrix $T_{s_j}$ so that the $\mathbf{v}_2$ matches the transformed filter $T_{s_j}^T \mathbf{w}_j$, i.e., $\mathbf{v}_1^T \mathbf{w}_j \approx \mathbf{v}_2^T T_{s_j}^T \mathbf{w}_j$.[4] Therefore, $\mathbf{v}_1$ and $\mathbf{v}_2$ will both activate $z_j$ after probabilistic max pooling. Figure 1 illustrates this idea.

---

[1] We will describe extensions to other feature learning algorithms in Section 3.4.

[2] Due to space constraints, we present only the case of binary-valued input variables; however, the RBM with real-valued input variables can be formulated straightforwardly (Hinton & Salakhutdinov, 2006; Lee et al., 2008).

[3] A similar technique is used in convolutional deep belief networks (Lee et al., 2011), in which spatial probabilistic max pooling is applied over a small spatial region.

[4] Note that the transpose $T_s^T$ of a transformation ma-

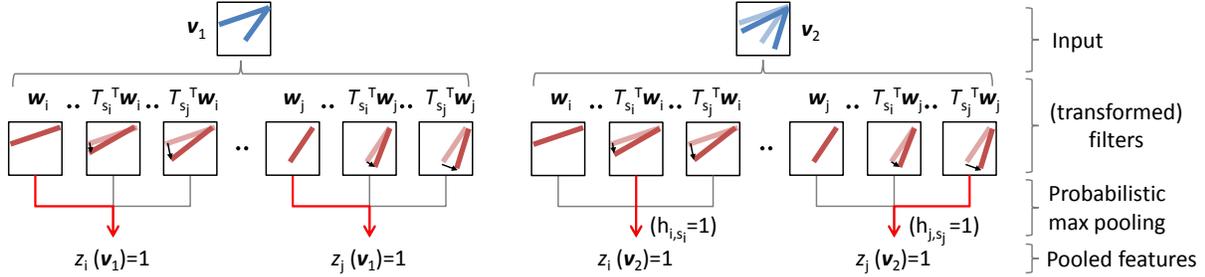

*Figure 1.* Feature encoding of TIRBM. Here, $\mathbf{v}_1$ and $\mathbf{v}_2$ denote two image patches, and the shaded pattern inside $\mathbf{v}_2$ reflects $\mathbf{v}_1$. The shaded patterns in transformed filters show the corresponding original filters $\mathbf{w}_i$ or $\mathbf{w}_j$. The filters selected via probabilistic max pooling across the set of transformations are shown in red arrows (e.g., in the rightmost example, the hidden unit $h_{j,s_j}$ corresponding to the transformation $T_{s_j}$ and the filter $\mathbf{w}_j$ contributes to activate $z_j(\mathbf{v}_2)$.) In this illustration, we assumed $D_1 = D_2$ and also the existence of the *identity* transformation.

Compared to the regular RBM, the TIRBM can learn more diverse patterns, while keeping the number of parameters small. Specifically, multiplying transformation matrix (e.g., $T_s^T \mathbf{w}_j$) can be viewed as increasing the number of filters by the factor of $S$, but without significantly increasing the number of parameters due to parameter sharing. In addition, by pooling over local transformations, the filters can learn invariant representations (i.e., $z_j$'s) to these transformations.

The conditional probabilities are computed as follows:

$$p(h_{j,s} = 1|\mathbf{v}) = \frac{\exp\left(\mathbf{w}_j^T T_s \mathbf{v} + b_{j,s}\right)}{1 + \sum_{s'} \exp\left(\mathbf{w}_j^T T_{s'} \mathbf{v} + b_{j,s'}\right)} \quad (7)$$

$$p(v_i = 1|\mathbf{h}) = \text{sigmoid}(\sum_{j,s}(T_s^T \mathbf{w}_j)_i h_{j,s} + c_i) \quad (8)$$

Similar to RBM training, we use stochastic gradient descent to train TIRBM. The gradient of the log-likelihood is approximated via contrastive divergence by taking the gradient of the energy function (Equation (5)) with respect to the model parameters.

### 3.2. Sparse TIRBM

The sparseness of the feature representation is often a desirable property. By following Lee et al.'s (2008) approach, we can extend our model to sparse TIRBM by adding the following regularizer for a given set of data $\{\mathbf{v}^{(1)}, \cdots, \mathbf{v}^{(N)}\}$ to the negative log-likelihood:

$$\mathcal{L}_{\text{sp}} = \sum_{j=1}^{K} \mathcal{D}\left(p, \frac{1}{N} \sum_{n=1}^{N} \mathbb{E}[z_j^{(n)}|\mathbf{v}^{(n)}]\right) \quad (9)$$

where $\mathcal{D}$ is a distance function; $p$ is the target sparsity.

The expectation of pooled activation is written as

$$\mathbb{E}[z_j|\mathbf{v}] = \frac{\sum_s \exp\left(\mathbf{w}_j^T T_s \mathbf{v} + b_{j,s}\right)}{1 + \sum_s \exp\left(\mathbf{w}_j^T T_s \mathbf{v} + b_{j,s}\right)}. \quad (10)$$

Note that we regularize over the pooled hidden units $z_j$ rather than individual hidden units $h_{j,s}$. In our experiments, we used L2 distance for $\mathcal{D}(\cdot,\cdot)$, but one can also use KL divergence for the sparsity penalty.

### 3.3. Generating transformation matrices

In this section, we discuss how to design the transformation matrix $T$. For the ease of presentation, we assume 1-d transformations, but it can be extended to 2-d cases (e.g., image transformations) straightforwardly. Further, we assume the case of $D_1 = D_2 = D$ here; but, we will discuss general cases later.

As mentioned previously, $T \in \mathbb{R}^{D \times D}$ is a linear transformation matrix from $\mathbf{x} \in \mathbb{R}^D$ to $\mathbf{y} \in \mathbb{R}^D$; i.e., each coordinate of $\mathbf{y}$ is constructed via linear combination of the coordinates in $\mathbf{x}$ with weight matrix $T$ as follows:

$$y_i = \sum_{j=1}^{D} T_{ij} x_j, \forall i = 1, \cdots, D. \quad (11)$$

For example, shifting by $s$ can be defined as

$$T_{ij}(s) = \begin{cases} 1 & \text{if } i = j + s, \\ 0 & \text{otherwise.} \end{cases}$$

For 2-d image transformations such as rotation and scaling, the contribution of input coordinates to each output coordinate is computed with bilinear interpolation. Since $T$'s are pre-computed once and usually sparse, Equation (11) can be computed efficiently.

### 3.4. Extensions to other methods

We emphasize that our transformation-invariant feature learning framework is not limited to the energy-based probabilistic models, but can be extended to other unsupervised learning methods as well.

---

trix $T_s$ also induces a linear transformation. In the paper, we will occasionally abuse the term "transformation" to denote these two cases, as long as the context is clear.

First, it can be readily adapted to autoencoders by defining the following softmax encoding and sigmoid decoding functions:

$$f_{j,s}(\mathbf{v}) = \frac{\exp\left(\mathbf{w}_j^T T_s \mathbf{v} + b_{j,s}\right)}{1 + \sum_{s'} \exp\left(\mathbf{w}_j^T T_{s'} \mathbf{v} + b_{j,s'}\right)} \quad (12)$$

$$\hat{v}_i = \text{sigmoid}(\sum_{j,s}(T_s^T \mathbf{w}_j)_i f_{j,s} + c_i) \quad (13)$$

Following the idea of TIRBM, we can also formulate the transformation-invariant sparse coding as follows:

$$\min_{\mathbf{W},\mathbf{H}} \sum_n \|\sum_{j=1}^K \sum_{s=1}^S T_s^T \mathbf{w}_j h_{j,s}^{(n)} - \mathbf{v}^{(n)}\|^2, \quad (14)$$

$$\text{s.t. } \|\mathbf{H}\|_0 \leq \gamma, \|\mathbf{H}(j,:)\|_0 \leq 1, \|\mathbf{w}_j\|_2 \leq 1, \quad (15)$$

where $\gamma$ is a constant. The second constraint in (15) can be understood as an analogy to the softmax constraint in Equation (6) of TIRBMs.

Similar to standard sparse coding, we can optimize the parameters by alternately optimizing $\mathbf{W}$ and $\mathbf{H}$ while fixing the other. Specifically, $\mathbf{H}$ can be (approximately) solved using Orthogonal Matching Pursuit, and therefore we refer this algorithm a transformation-invariant Orthogonal Matching Pursuit (TIOMP).

## 4. Related Work

Researchers have made significant efforts to develop invariant feature representations. For example, the rotation- or scale-invariant descriptors, such as SIFT (Lowe, 1999), have shown a great success in many computer vision applications. However, these image descriptors usually demand a domain-specific knowledge with a significant amount of hand-crafting.

As an alternative approach, several unsupervised learning algorithms have been proposed to learn robust feature representations automatically from the sensory data. As an example, the denoising autoencoder (Vincent et al., 2008) can learn robust features by trying to reconstruct the original data from the hidden representations of randomly perturbed data generated from the distortion processes, such as adding noise or multiplying zeros for randomly selected coordinates.

Among types of transformations relating to temporally or spatially correlated data, translation has been extensively studied in the context of unsupervised learning. Specifically, convolutional training (LeCun et al., 1989; Kavukcuoglu et al., 2010; Zeiler et al., 2010; Lee et al., 2011; Sohn et al., 2011) is one of the most popular methods that encourages shift-invariance during the feature learning. For example, the convolutional deep belief network (CDBN) (Lee et al., 2011), which is composed of multiple layers of convolutional restricted Boltzmann machines and probabilistic max pooling, can learn a representation invariant to local translation.

Besides translation, however, learning invariant features for other types of image transformations have not been extensively studied. In contemporary work of ours, Kivinen and Williams (2011) proposed the transformation equivariant Boltzmann machine, which shares a similar mathematical formulation to our models in that both try to infer the best matching filters by transforming them using linear transformation matrices. However, while their model was motivated from the "global equivariance", the main purpose of our work is to learn locally-invariant features that can be useful in classification tasks. Thus, rather than considering an algebraic group of transformation matrices (e.g., "full rotations"), we focus on the variety of local transformations that include rotation, translation as well as scale variations. Furthermore, we effectively address the boundary effects that can be highly problematic in scaling and translation operators by forming a non-square $T$ matrix, rather than zero-padding.[5] In addition, we present a general framework of transformation-invariant feature learning and show extensions based on the autoencoder and sparse coding. Overall, our argument is strongly supported by the state-of-the-art performance in image and audio classification tasks.

As another related work, feature learning methods with topographic maps can also learn invariant representations (Hyvärinen et al., 2001; Kavukcuoglu et al., 2009). Compared to these methods, our model is more compact and has fewer parameters to train since it *factors out* the (filter) weights and their transformations. In addition, given the same number of parameters, our model can represent more diverse and variable input patterns than topographic filter maps.

## 5. Experiments

We begin by describing the notation. For images, we assume a receptive field size of $r \times r$ pixels (for input image patches) and a filter size of $w \times w$ pixels. We define $gs$ to denote the number of pixels corresponding to the transformation (e.g., translation or scaling). For example, we translate the $w \times w$ filter across the $r \times r$ receptive field with a stride of $gs$ pixels (Figure 2(a)), or scale down from $(r-l \cdot gs) \times (r-l \cdot gs)$ to $w \times w$ (where $0 \leq l \leq \lfloor (r-w)/gs \rfloor$) by sharing the same center for the filter and the receptive field (Figure 2(b)).

---

[5] We observed that learning with zero-padded squared transformation matrices showed significant boundary effect in its visualization of filters, and this often resulted in significantly worse classification performance.

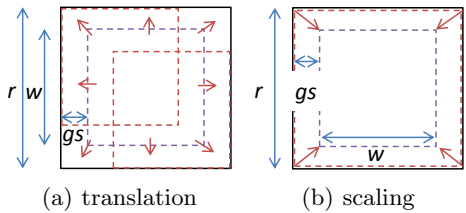

(a) translation    (b) scaling

*Figure 2.* Translation and scale transformations on images.

For classification tasks, we used the posterior probability of the pooled hidden unit (Equation (10)) as a feature. Note that the dimension of the extracted feature vector for each image patch is $K$, not $K \times S$. Thus, we argue that the performance gain of the TIRBM over the regular RBM comes from the better representation (i.e., transformation-invariant features), rather than from the classifier's use of higher-dimensional features.

### 5.1. Handwritten digit recognition with prior transformation information

First, we verified the performance of our algorithm on the variations of a handwritten digit dataset, assuming that the transformation information is given. From the MNIST variation datasets (Larochelle et al., 2007), we tested on "mnist-rot" (rotated digits, referred to as *rot*) and "mnist-rot-back-image" (rotated digits with background images, referred to as *rot-bgimg*). To further evaluate with different types of transformations, we created four additional datasets that contain scale and translation variations with and without random background (referred to as *scale*, *scale-bgrand*, *trans*, and *trans-bgrand*, respectively).[6] Some examples are shown in Figure 3.

For these datasets, we trained sparse TIRBMs on the image patches of size $28 \times 28$ pixels with data-specific transformations. For example, we considered 16 equally-spaced rotations (i.e., the step size of $\frac{\pi}{8}$) for the *rot* and *rot-bgimg* datasets. Similarly, for the *scale* and *scale-bgrand* datasets, we generated scale-transformation matrices with $w = 20$ and $gs = 2$, which can map from $(28 - 2l) \times (28 - 2l)$ pixels to $20 \times 20$ pixels with $l \in \{0, ..., 4\}$. For the *trans* and *trans-bgrand* datasets, we set $w = 24$ and $gs = 2$ to have total nine translation matrices that cover the $28 \times 28$ regions using $24 \times 24$ pixels with a horizontal and vertical stride of 2 pixels. For classification, we trained 1,000 filters for both sparse RBMs and sparse TIRBMs and used a softmax classifier. We used 10,000 examples for the training set, 2,000 examples for the validation set, and 50,000 examples for the test set.

[6] We followed the generation process described in http://www.iro.umontreal.ca/~lisa/twiki/bin/view.cgi/Public/MnistVariations to create the customized scaled and translated digits. For example, we randomly selected the scale-level uniformly from 0.3 to 1 or the number of pixel shifts in horizontal and vertical directions without making any truncation of the foreground pixels. For the datasets with random background, we randomly added the uniform noise in [0, 1] to the background pixels.

*Table 1.* Test classification error on MNIST transformation datasets. The best-performing methods for each dataset are shown in bold.

| Dataset | RBM | TIRBM | Transformation |
|---|---|---|---|
| rot | 15.6% | **4.2%** | rotation |
| rot-bgimg | 54.0% | **35.5%** | rotation |
| scale | 6.1% | **5.5%** | scaling |
| scale-bgrand | 32.1% | **23.7%** | scaling |
| trans | 15.3% | **9.1%** | translation |
| trans-bgrand | 57.3% | **43.7%** | translation |

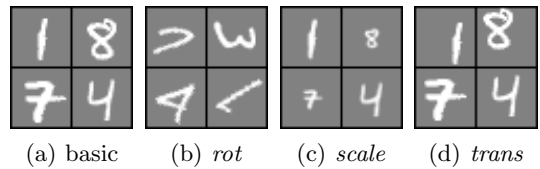

(a) basic  (b) *rot*  (c) *scale*  (d) *trans*

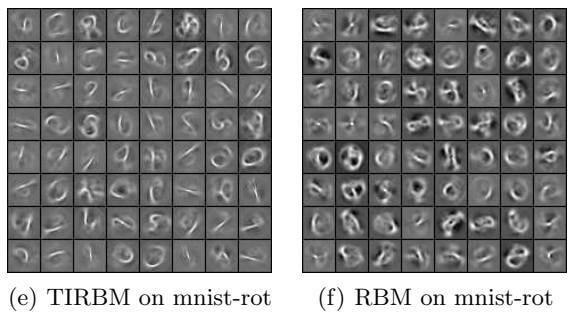

(e) TIRBM on mnist-rot    (f) RBM on mnist-rot

*Figure 3.* (top) Samples from the handwritten digit datasets with (a) no transformations, (b) rotation, (c) scaling, and (d) translation. (bottom) Learned filters from mnist-rot dataset with (e) the sparse TIRBM and (f) the sparse RBM, respectively.

As reported in Table 1, our method (sparse TIRBMs) consistently outperformed the baseline method (sparse RBMs) for all datasets. These results suggest that the TIRBMs can learn better representations for the foreground objects by transforming the filters. It is worth noting that our error rates for the mnist-rot and mnist-rot-back-image datasets are also significantly lower than the best published results obtained with stacked denoising autoencoders (Vincent et al., 2010) (9.53% and 43.75%, respectively).

For qualitative evaluation, we visualize the learned filters on the mnist-rot dataset trained with the sparse TIRBM (Figure 3(e)) and the sparse RBM (Figure 3(f)), respectively. The filters learned from sparse TIRBMs show much clearer pen-strokes than those learned from sparse RBMs, which partially explains

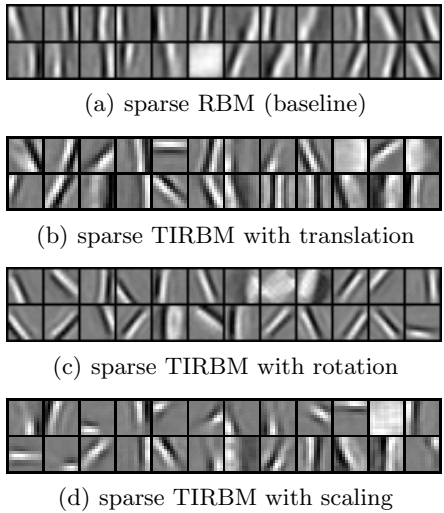

(a) sparse RBM (baseline)

(b) sparse TIRBM with translation

(c) sparse TIRBM with rotation

(d) sparse TIRBM with scaling

*Figure 4.* Visualization of filters trained with RBM and TIRBMs on natural images. We trained 24 filters and used nine translations with a step size of 1 pixel, five rotations with a step size of $\pi/8$ radian, and two-level scale transformations with a step size ($gs$) of 2 pixels, respectively.

the impressive classification performance.

### 5.2. Learning invariant features from natural images

For handwritten digit recognition in the previous section, we assumed prior information on *global* transformations on the image (e.g., translation, rotation, and scale variations) for each dataset. This assumption enabled the proposed TIRBMs to achieve significantly better classification performance than the baseline method, since the data-specific transformation information was encoded in the TIRBM.

However, for natural images, it is not reasonable to assume such *global* transformations due to the complex image structures. In fact, recent literature (Yu et al., 2011; Lee et al., 2011; Vincent et al., 2008) suggests that some level of invariance to *local* transformations (e.g., few pixel translation or coordinate-wise noise) leads to improved performance in classification. From this viewpoint, it makes more sense to learn representations with local receptive fields that are invariant to *generic* image transformations (e.g., small amounts of translation, rotation, and scaling), which does not require data-specific prior information.

We visualize the learned TIRBM filters in Figure 4, where we used the 14×14 natural image patches taken from the van Hateren dataset (van Hateren & van der Schaaf, 1998). The baseline model (sparse RBM) learns many similar vertical edges (Figure 4(a)) that are shifted by a few pixels, whereas our methods can learn diverse patterns, including diagonal and horizontal edges, as shown in Figure 4(b), 4(c), and 4(d).

*Table 2.* Test classification accuracy on CIFAR-10 dataset. 1,600 filters were used unless otherwise stated. The numbers with † and ‡ are from (Coates et al., 2011) and (Coates & Ng, 2011a), respectively.

| Methods (Linear SVM) | Accuracy |
|---|---|
| sparse RBM† (baseline) | 72.4% |
| sparse TIRBM (scaling only) | 76.8% |
| sparse TIRBM (rotation only) | 77.7% |
| sparse TIRBM (translation only) | 77.5% |
| sparse TIRBM (combined) | 78.8% |
| sparse TIRBM (combined, $K$= 4,000) | 80.1% |
| TIOMP-1/T (combined) | 80.7% |
| TIOMP-1/T (combined, $K$= 4,000) | **82.2%** |
| VQ ($K$= 4,000)† | 79.6% |
| OMP-1/T ($K$= 1,600)‡ | 79.4% |
| OMP-1/T ($K$= 6,000)‡ | 81.5% |
| convolutional DBN (Krizhevsky, 2010) | 78.9% |
| deep NN (Ciresan et al., 2011) | 80.5% |
| deep NN (Coates & Ng, 2011b) | 82.0% |

*Table 3.* Test classification accuracy on STL-10. 1,600 filters were used for all experiments.

| Methods (Linear SVM) | Acc.± std. |
|---|---|
| sparse RBM | 55.0 ± 0.5% |
| sparse TIRBM (scaling only) | 55.9 ± 0.7% |
| sparse TIRBM (rotation only) | 57.0 ± 0.7% |
| sparse TIRBM (translation only) | 57.8 ± 0.5% |
| sparse TIRBM (combined) | 58.7 ± 0.5% |
| VQ ($K$= 1,600) (Coates & Ng, 2011a) | 54.9 ± 0.4% |
| SC ($K$= 1,600) (Coates & Ng, 2011a) | 59.0 ± 0.8% |
| deep NN (Coates & Ng, 2011b) | **60.1 ± 1.0%** |

These results suggest that TIRBMs can learn diverse sets of filters, which is reminiscent of the effects of convolutional training (Kavukcuoglu et al., 2010). However, our model is much easier to train than convolutional models, and it can further handle generic transformations beyond translations.

### 5.3. Object recognition

We evaluated on image classification tasks using two datasets. First, we tested on the widely used CIFAR-10 dataset (Krizhevsky, 2009), which is composed of 50,000 training and 10,000 testing examples with 10 categories. Rather than learning features from the whole image (32 × 32 pixels), we trained TIRBMs on local image patches while keeping the RGB channels. As suggested by Coates et al.(2011), we used the fixed filter size $w = 6$ and determined the receptive field size depending on the types of transformations.[7] Then, after unsupervised training with TIRBM, we used the convolutional feature extraction scheme, also following the Coates et al.(2011). Specifically, we computed the TIRBM pooling-unit activations for each local $r \times r$ pixel patch that was densely extracted with a stride of 1 pixel, and averaged the patch-level activations over each of the 4 quadrants in the image. Eventually, this procedure yielded $4K$-dimensional feature vectors for each image, which were fed into an L2-regularized linear SVM. We performed 5-fold cross validation to determine the hyperparameter $C$.

---

[7]For example, we used $r = 6$ for rotations. For both scale variations or translations, we used $r = 8$ and $gs = 2$.

For comparison to the baseline model, we separately evaluated the sparse TIRBMs with a single type of transformation (translation, rotation, or scaling) using $K = 1,600$. As shown in Table 2, each single type of transformation in TIRBMs brought a significant performance gain over the baseline sparse RBMs. The classification performance was further improved by combining different types of transformations into a single model.

In addition, we also report the classification results obtained using TIOMP-1 (see Section 3.4) for unsupervised training. In this experiment, we used the following two-sided soft thresholding encoding function:

$$f_j = \max_s \left\{ \max(\mathbf{w}_j^T T_s \mathbf{v} - \alpha, 0) \right\}$$
$$f_{j+K} = \max_s \left\{ \max(-\mathbf{w}_j^T T_s \mathbf{v} - \alpha, 0) \right\},$$

where $\alpha$ is a constant threshold that was cross-validated. As a result, we observed about 1% improvement over the baseline method (OMP-1/T) using 1,600 filters, which supports the argument that our transformation-invariant feature learning framework can be effectively transferred to other unsupervised learning methods. Finally, by increasing the number of filters ($K = 4,000$), we obtained better results (82.2%) than the previously published results using single-layer models, as well as those using deep networks.

We also performed the object classification task on STL-10 dataset (Coates et al., 2011), which is more challenging due to the smaller number of labeled training examples (100 per class for each training fold). Since the original images are 96×96 pixels, we downsampled the images into 32×32 pixels, while keeping the RGB channels. We followed the same unsupervised training and classification pipeline as we did for CIFAR-10. As reported in Table 3, there were consistent improvements in classification accuracy by incorporating the various transformations in learning algorithms. Finally, we achieved 58.7% accuracy using 1,600 filters, which is competitive to the best published single layer result (59.0%).

### 5.4. Phone classification

To show the broad applicability of our method to other data types, we report the 39-way phone classification accuracy on the TIMIT dataset. By following (Ngiam et al., 2011), we generated 39-dimensional MFCC features and used 11 contiguous frames of them as an input patch. For TIRBMs, we applied three temporal translations with the stride of 1 frame.

Table 4. Phone classification accuracy on the TIMIT core test set using linear SVMs.

| Methods (Linear SVM) | Accuracy |
|---|---|
| MFCC | 68.2% |
| sparse RBM | 76.3% |
| sparse TIRBM | **77.6%** |
| sparse coding (Ngiam et al., 2011) | 76.8% |
| sparse filtering (Ngiam et al., 2011) | 75.7% |

Table 5. Phone classification accuracy on the TIMIT core test set using RBF-kernel SVMs.

| Methods (RBF-kernel SVM) | Accuracy |
|---|---|
| MFCC (baseline) | 80.0% |
| MFCC + TIRBM ($K$= 256) | 81.0% |
| MFCC + TIRBM ($K$= 512) | **81.5%** |
| MFCC + SC (Ngiam et al., 2011) | 80.1% |
| MFCC + SF (Ngiam et al., 2011) | 80.5% |
| MFCC + CDBN (Lee et al., 2009) | 80.3% |
| H-LMGMM (Chang & Glass, 2007) | 81.3% |

First, we compared the classification accuracy using the linear SVM to evaluate the performance gain coming from the unsupervised learning algorithms, by following the experimental setup in (Ngiam et al., 2011).[8] As reported in Table 4, the TIRBM showed an improvement over the sparse RBM, as well as the sparse coding and sparse filtering.

In the next setting, we used the RBF-kernel SVM (Chang & Lin, 2011) on the extracted features that are concatenated with MFCC features. We used default RBF kernel width for all experiments and performed cross-validation on the C values. As shown in Table 5, combining MFCC with TIRBM features was the most effective and resulted in 1% improvement in classification accuracy over the baseline MFCC features. By increasing the number of TIRBM features to 512, we were able to beat the best published results on the TIMIT phone classification tasks using hierarchical LM-GMM classifier (Chang & Glass, 2007).

## 6. Conclusion and Future Work

In this work, we proposed novel feature learning algorithms that can achieve invariance to the set of predefined transformations. Our method can handle general transformations (e.g., translation, rotation, and scaling), and we experimentally showed that learning invariant features for such transformations leads to strong classification performance. In future work, we plan to work on learning transformations from the data and combine it with our algorithm. By automatically learning transformation matrices from the data, we will be able to learn more robust features, which will potentially lead to significantly better feature representations.

---

[8] We used $K = 256$ with a two-sided encoding function by using the positive and negative weight matrices $[\mathbf{W}, -\mathbf{W}]$, as suggested by Ngiam et al. (2011).

### Acknowledgments

This work was supported in part by a Google Faculty Research Award.